\documentclass[lettersize,journal]{IEEEtran}
\usepackage{amsmath,amsfonts}
\usepackage{algorithmic}
\usepackage{array}
\usepackage[caption=false,font=normalsize,labelfont=sf,textfont=sf]{subfig}
\usepackage{textcomp}
\usepackage{stfloats}
\usepackage{url}
\usepackage{verbatim}
\usepackage{graphicx}
\usepackage{hyperref}
\usepackage{siunitx}

\def\BibTeX{{\rm B\kern-.05em{\sc i\kern-.025em b}\kern-.08em
    T\kern-.1667em\lower.7ex\hbox{E}\kern-.125emX}}
\usepackage{balance}
\begin{document}
\title{AERIAL-CORE: AI-Powered Aerial Robots for Inspection and Maintenance of Electrical Power Infrastructures}
\author{A. Ollero, Fellow, IEEE, A. Suarez, C. Papaioannidis, I. Pitas, Life Fellow, IEEE, J.M. Marredo, V. Duong, student member, IEEE, E. Ebeid, Senior member, IEEE, V. Krátký, M. Saska, C. Hanoune, A. Afifi, A. Franchi, Fellow, IEEE, C. Vourtsis, D. Floreano, Fellow, IEEE, G. Vasiljevic, member, IEEE, S. Bogdan, Senior member, IEEE, A. Caballero, F. Ruggiero, Senior member, IEEE, V. Lippiello, Senior member, IEEE, C. Matilla, G. Cioffi, D. Scaramuzza, Senior member, IEEE,  J. R. Martinez-de Dios, B. C. Arrue, C. Martín, K. Zurad,  C. Gaitán, J. Rodriguez, A. Munoz, A. Viguria, Senior member, IEEE
\thanks{This work was supported by the European Union’s Horizon 2020 Research and Innovation Program under grant agreement no. 871479 (AERIAL-CORE).}}

\maketitle

\begin{abstract} 
Large-scale infrastructures are prone to deterioration due to age, environmental influences, and heavy usage. Ensuring their safety through regular inspections and maintenance is crucial to prevent incidents that can significantly affect public safety and the environment. This is especially pertinent in the context of electrical power networks, which, while essential for energy provision, can also be sources of forest fires. 
Intelligent drones have the potential to revolutionize inspection and maintenance, eliminating the risks for human operators, increasing productivity, reducing inspection time, and improving data collection quality. However, most of the current methods and technologies in aerial robotics have been trialed primarily in indoor testbeds or outdoor settings under strictly controlled conditions, always within the line of sight of human operators. Additionally, these methods and technologies have typically been evaluated in isolation, lacking comprehensive integration. This paper introduces the first autonomous system that combines various innovative aerial robots. This system is designed for extended-range inspections beyond the visual line of sight, features aerial manipulators for maintenance tasks, and includes support mechanisms for human operators working at elevated heights. The paper further discusses the successful validation of this system on numerous electrical power lines, with aerial robots executing flights over 10 kilometers away from their ground control stations.
\end{abstract}


\section*{Multimedia}
A video summarizing the AERIAL-CORE project can be found at \url{https://www.youtube.com/watch?v=Oyw7VwM7sCs}. 

\section{Introduction}
\label{sec:Intro}

\IEEEPARstart{E}{very} year, many accidents occur due to the failure of civil and industrial infrastructures. Their inspection and maintenance (I\&M) are mandatory to avoid these risks.  I\&M of large infrastructures costs many billions every year, and, in many cases, the risk for human workers is also very high. Thus, work at elevated heights is the first cause of fatal labor accidents in many countries. Electrical power grids are a critical infrastructure with tens of millions of km (100 million have been estimated worldwide and 5 million km only in the European Union) of power lines connecting hundreds of millions of customers. Frequent I\&M is mandatory to avoid power outages to many people and environmental impact such as forest fires due to the contact of trees to the lines (California’s second-largest wildfire was sparked when power lines came in contact with a tree\footnote {\url{https://edition.cnn.com/2022/01/05/us/dixie-fire-power-lines-cause-pge}}) and also the collision and electrocution of birds, which can also cause power cuts, power outages, and start fires during periods of drought. Thus, in some countries, installing bird diverters reflecting the solar light to avoid bird collisions is mandatory.  

Conventional inspection is very costly (e.g., inspection using manned helicopters costs 150 €/km). The maintenance activities are even more expensive and very risky for humans. The total inspection and maintenance cost in Europe has been estimated at more than 2.2 billion €/year. Climate change is increasing the rate of natural disasters and more adverse weather conditions that usually damage electrical towers and cables. This fact demands the development of new technologies for their management in reducing the I\&M operation time and costs. 
Most market studies of drones point to an exponential increase in the application for I\&M \footnote {\url {https://ec.europa.eu/commission/presscorner/detail/en/}}.
However, up to now, inspection drones have remote human pilots and operate mainly within the Visual Line of Sight (VLOS). Autonomous aerial systems for inspection are very few, with very constrained endurance/range of flight, and are subject to environmental constraints, such as wind gusts, which deteriorate the inspection quality. The situation is even worse in aerial manipulation. Most aerial robotic manipulation systems have only been demonstrated in indoor testbeds or outdoors in very controlled conditions (see review in \cite{ollero2021past}). Aerial and rolling cantilever robots have been applied to I\&M of power lines, usually within VLOS and with different degrees of autonomy (see \cite{chen2023} and its references).

The main contribution of this paper is the presentation, for the first time, of different fully autonomous innovative aerial robotics technologies and the integrated system for I\&M of electrical power lines, including autonomous inspection Beyond Visual Line of Sight (BVLOS). This system has been developed in the framework of the European Union (EU)-funded AERIAL-CORE project \footnote{\url{https://aerial-core.eu/}}.  
The BVLOS autonomous inspection is executed through the use of innovative morphing platforms capable of hovering for detailed inspection and flying over long distances. It employs advanced object-tracking methods and environmental mapping around the power lines, facilitating the measurement of the distance between vegetation and conductors. Additionally, the system incorporates multi-robot teaming for extended-range inspections, including an autonomous battery charging feature. This enables the comprehensive inspection of vast electrical power grids with minimal human intervention. 

The maintenance aspect of aerial robotic manipulation is enhanced by the novel approach of perching on electrical conductors. This technique significantly improves precision and the amount of force applied compared to existing aerial manipulation methods. Another innovative feature is the capability to recharge drone batteries using the energy from the electrical lines themselves. Moreover, the system includes aerial robotic co-workers designed to aid human operators working at elevated heights. These robotic co-workers are equipped with autonomous gesture recognition, can transport tools, and can perform voltage checks.

The subsequent section of the paper details the AERIAL-CORE mission and its validation scenarios. The three following sections are dedicated to discussing the subsystems: long-range inspection, manipulation, and co-working. The paper then presents the integrated system as a whole. It concludes with a section on the overall conclusions drawn from the research and potential future applications of these technologies.

\section{Mission and Scenario}
\label{sec:MissinAndScenario}

The mission of the AERIAL-CORE EU project was to demonstrate I\&M in the medium voltage power lines located in the aerial space of the ATLAS experimental flight center at Villacarrillo (Spain), which has a segregated airspace reserve with an approximate extension of more than \SI{1000}{\kilo\meter\squared}.

Meeting drone regulations is the main bottleneck for developing the kind of applications presented in this paper. Then, strong efforts were devoted to obtaining the authorizations fulfilling the European drone regulation \footnote{\url{https://www.easa.europa.eu/en/domains/civil-drones}} in ATLAS and regular industrial operations.  Particularly, the long-range inspection was performed under the Specific Category. The UAVs performing these flights operated under two different regulatory conditions: (1) Operational authorization, which grants the tilt-rotor aircraft presented in the next section to fly distances of up to \SI{10}{\kilo\meter} while inspecting power lines, and (2) Standard STS-ES-02 Scenario, which enables multi-rotor flights to be carried out up to \SI{2}{\kilo\meter} from the ground control station with observers and maintaining an altitude below \SI{120}{\meter} above ground level.

The aerial manipulators in Section IV are also of two types. The DAP-C type for installing bird diverters (see IV-A) is within the Open Category since the Maximum Take-Off Weight (MTOW) of the aerial system is below \SI{25}{\kilo\gram}, corresponding to Subcategory A3. On the other hand, the MLMP robot(see IV-B) weighs \SI{40}{\kilo\gram}, so it cannot fit into the Open Category, and was enclosed into the Specific Category (with operational authorization). 

Finally, in section V, the flights within the aerial co-working subsystem were all included in the Open Category.

\begin{figure*}[h!]
    \centering
    \includegraphics[width=1\textwidth]{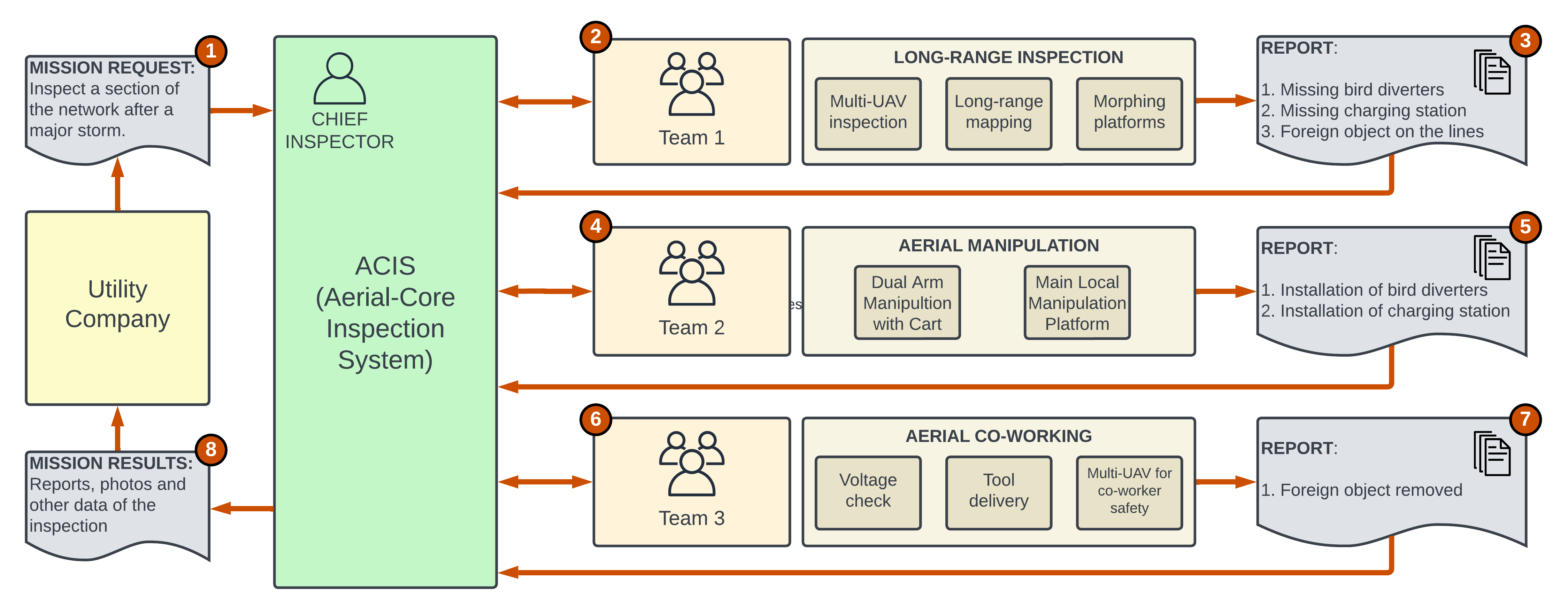}
    \caption{AERIAL-CORE intelligent aerial robotic system for inspection and maintenance of large infrastructures.}
    \label{fig:/Figures_Section_II/aerial-core_system}
\end{figure*}

The AERIAL-CORE mission was performed as shown in Figure \ref{fig:/Figures_Section_II/aerial-core_system}. The utility company was first tasked to inspect the network after a major storm. Then, the chief inspector launched a long-range inspection (see Section III). The results are in the report file, including missing bird diverters, a missing charging station, and a foreign object hanging from the line. Then, manipulation was launched to carry out the installation of bird diverters and the charging station, which was done using two different manipulation systems (see Figure \ref{fig:/Figures_Section_II/aerial-core_system} and Section IV), and a report is generated. Finally, it  started co-working, involving voltage checks, tool delivery, and the multi-UAV-based monitoring of the safety of the human co-worker removing the foreign object (detailed in Section V)

\section{Long-Range Inspection}
\label{sec:LongRange}

\subsection{Morphing for Long-Range Inspection}
\label{sec::MorphingLongRange}

AERIAL-CORE has considered different morphing schemes,  including fixed wing-rotary wing morphing aerial vehicles, morphing multi-rotors with morphing between omnidirectional and conventional configurations, and fixed wing-flapping wing hybrid vehicles. In the following, we summarize two platforms with morphing for long-range inspection. 

\begin{figure}[h!]
    \centering
    \includegraphics[width=1\columnwidth]{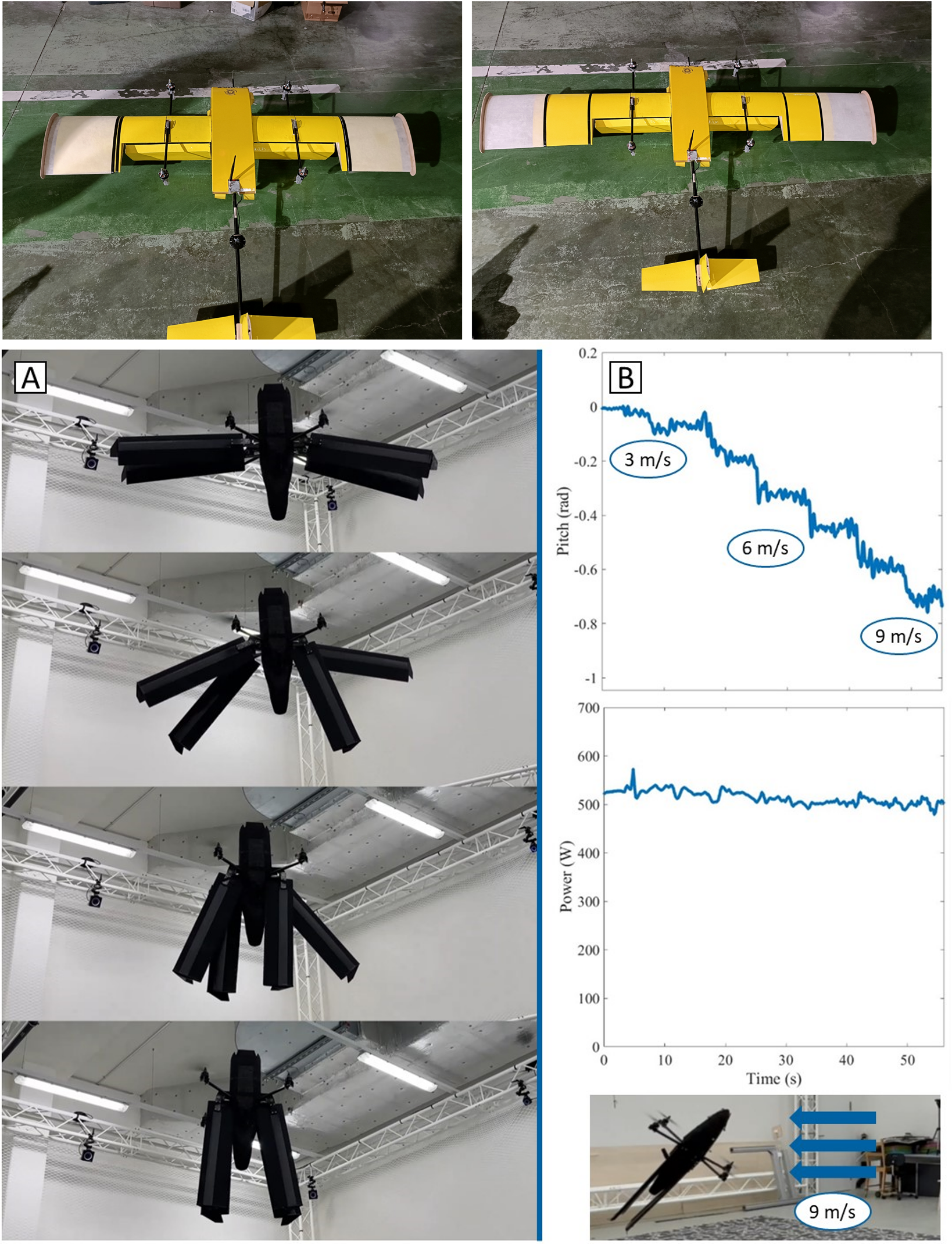}
    \caption{\textbf{(Top)} MARVIN-5-M. \textbf{(Bottom)} (A) Morpho in hovering flight with different wing configurations. (B) Morpho in hovering experiments. Demonstration of the pitch angle related to wind speed and energy consumption.}
    \label{fig:/Figures_Section_III/MorphingAll}
\end{figure}

Marvin-5 is a UAV that can fly as a fixed wing and change to the copter mode, varying the orientation of its four rotors for Vertical Take-off and Landing (VTOL). When flying as a fixed wing, the minimum aircraft’s stall speed is 13.30 m/s and the total ground speed is over 20 m/s. These high speeds generate defects in the data capture process, such as motion blur in the images captured by the visual and thermal cameras, as well as low-density point clouds generated by the LIDAR sensor. To solve these issues, the natural solution would be to increase the wing size to reduce the stall speed. However, the bigger the wing, the more sensible the aircraft is in copter mode to the wind. Then, the vehicle adapts its geometry to the different flight modes: smallest wingspan for copter mode (VTOL and hovering) and increased wingspan for optimal data capture plane mode. The system can adapt its wingspan autonomously during the flight to meet the optimal flight conditions. The system increases the aircraft's wingspan from 1.85 to 2.41 m by moving two empty elements with the same geometry as the rest of the wing. These empty elements are pushed by one servo on each side to achieve a wing surface of 0.709 m2 from 0.527 m2,  achieving a reduction of stall speed depending on the pitch of the aircraft. In addition to the VTOL, the wingspan can be adjusted dynamically during the flight to reduce consumption in the parts of the flight where no data is captured (connecting flights between power line segments). The resulting industrial aircraft with morphing capabilities has been called MARVIN-5-M (see Figure \ref{fig:/Figures_Section_III/MorphingAll} Top) and can have hundreds of operations without maintenance, 

In addition to the previously mentioned platform, a novel drone, called Morpho, is capable of performing both hovering and long-range horizontal flight was developed. It is a bioinspired, quad, morphing, biplane tailsitter VTOL that features four identical and independently actuated wings to combine the advantages of copters and fixed-wing in a single vehicle (Figure \ref{fig:/Figures_Section_III/MorphingAll} bottom). Morpho can transform into a stable and safe configuration with folded wings that will allow it to operate efficiently and precisely in vertical flight and while in proximity to humans and infrastructure. In horizontal flight, Morpho transforms into an efficient aerodynamic shape, enabling it to travel over several kilometers. The independent dynamic wing actuation improves the drone’s energy efficiency and maneuverability in hovering flight \cite{vourtsis_wind_2023}. The morphing capabilities allow Morpho to overcome the limitations of current commercially available state-of-the-art VTOL platforms with exposed wings and demonstrate energetic advantages in hovering flight. 
The efficient vertical flight configuration with retracted wings also sustains higher wind speeds than standard VTOLs with constantly exposed wings. This prototype has been tested up to wind speeds of 9 m/s while maintaining its hovering position and demonstrating stable or reduced energy consumption despite the gradual increase of the wind (Figure \ref{fig:/Figures_Section_III/MorphingAll} Bottom (B)).
Morpho is a 3.5 kg drone with a combined flight time of around 17 minutes. The drone's average speed in vertical flight is around 8 km/h, while it is about 60 km/h in horizontal flight. The drone is capable of autonomous waypoint navigation during flight. Morpho can transform from hovering to horizontal flight and vice versa. The complete flight envelope was demonstrated in ATLAS, where Morpho performed hovering and horizontal flights in a realistic mission scenario (video:\href{https://youtu.be/KfKCRcdeYYQ}{https://youtu.be/KfKCRcdeYYQ}).

\subsection{Tracking}
\label{sec::Tracking}

\begin{figure}[h!]
    \centering
    \includegraphics[width=1\columnwidth]{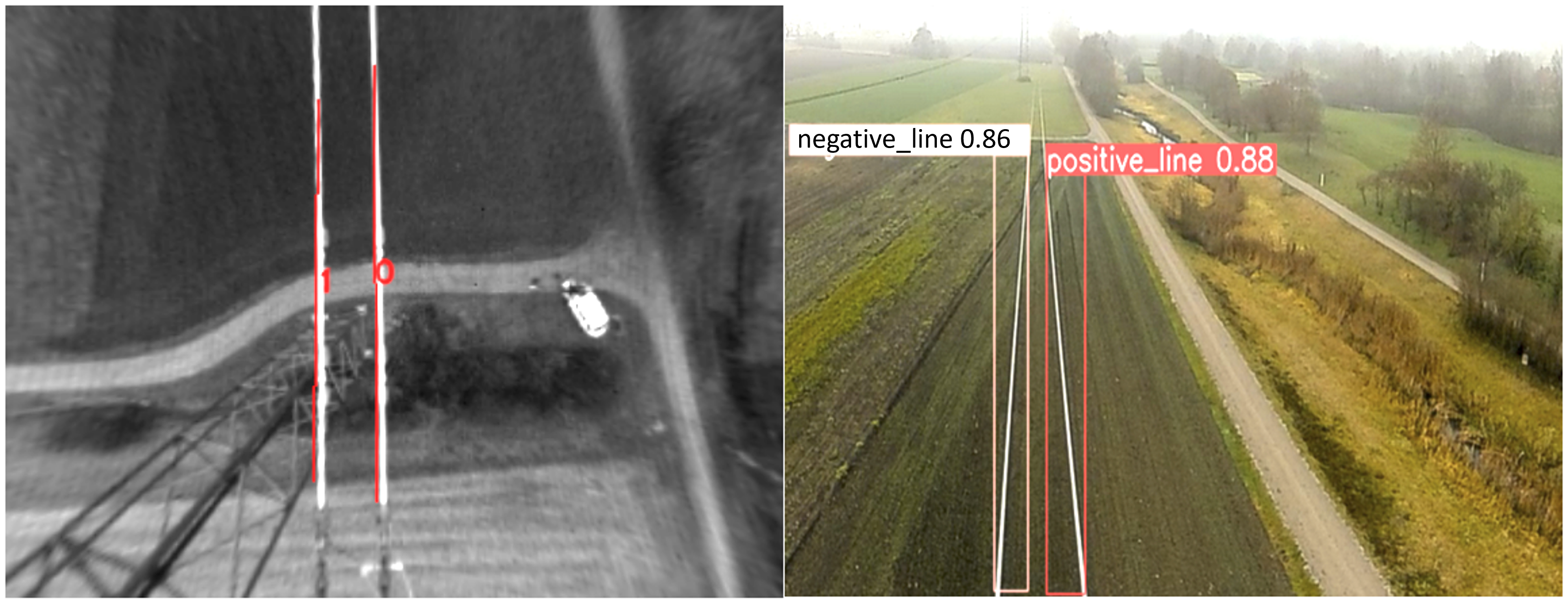}
    \caption{\textbf{(Left)} Output of the event-based tracker~\cite{dietsche2021powerline}. Our algorithm assigns to each tracked line a unique ID. \textbf{(Right)} Output of the RGB-based tracker~\cite{xing2023autonomous}. Our algorithm assigns to each tracked line its inclination (positive or negative), a confidence score, and a unique ID (not visualized).}
    \label{fig:UZH_tracking}
\end{figure}

In AERIAL-CORE, we have developed an RGB-based~\cite{xing2023autonomous} and an Event-based~\cite{dietsche2021powerline} power line tracker, (see Figure ~\ref{fig:UZH_tracking}).

The RGB-based power line tracker~\cite{xing2023autonomous} is a novel algorithm that extends the deep-learning–based object detector in~\cite{redmon2016you} to the case of power line detection. The perception module is trained only on synthetic data and transfers zero-shot to real-world images of power lines without fine-tuning. In this way, we overcome the problem of the limited amount of annotated data for supervised learning. The detector takes a single RGB image as input and outputs end points of the detected power lines in pixel coordinates. The center patch of each detection is matched with the prediction of the previous patch using the Hungarian method. We use a KLT tracker~\cite{lucas1981iterative} to perform tracking. To train our detector, we created a new synthetic dataset for power line detection based on a simulator.
Our model is trained on circa 30k simulated images. We show real-world deployment without fine-tuning on real images. 

Event cameras~\cite{Gallego20pami} are inherently robust to motion blur and have low latency and high dynamic range. Then, they are advantageous for the autonomous inspection of power lines with drones, where fast motions and challenging illumination conditions are ordinary. Our event-based power line tracker~\cite{dietsche2021powerline} identifies lines in the stream of events by detecting planes in the spatio-temporal signal and tracking them through time. Our algorithm can persistently track the power lines, with a mean lifetime of the line 10 times longer than existing approaches~\cite {everding2018low}. Both algorithms have been validated onboard a quadrotor equipped with an Nvidia Jetson TX2 ~\cite{foehn2022agilicious}.

\subsection{Long-range Mapping }
\label{sec::Mapping}

Mandatory power line inspections include the detection of violations of safety distances between the power line and electric towers and the surrounding vegetation. The traditional approach is based on capturing images and LIDAR data using manned helicopters and off-processing them in the office to obtain --weeks later-- the resulting segmented map, which is very inefficient and leads to repetitions of data gathering flights. In AERIAL-CORE, an online LIDAR-based 3D semantic mapping system for UAV long-range power-line inspection has been developed. The method is based on \emph{FAST-LIO2}, which was adapted by incorporating GNSS measurements to enable robustness to scenarios with a lack of geometrical features \cite{mapping1}. It was also enhanced with LIDAR-based segmentation that classifies the map point cloud into \emph{Powerlines}, \emph{Towers}, \emph{Vegetation}, and \emph{Soil} by combining LIDAR reflectivity region growing (to differentiate between metallic and non-metallic objects) and Principal Component Analysis (to consider the point-cloud spatial distribution) \cite{mapping1}.

The method was implemented onboard the \emph{LR-M} aerial robot developed by the Univ. of Seville. \emph{LR-M} is based on the DJI Matrice 600 platform (with BVLOS flight capabilities) and uses as the primary sensor a Livox Horizon 3D solid-state LIDAR, which provides non-repetitive scanning. A Jetson NVIDIA NX Xavier was used for onboard computation and logging. More than 80 experiments were conducted in BVLOS mission scenarios with different conditions and vegetation. Figure \ref{fig:multiUAV_inspection}-Bottom shows the map obtained online in the 3.7 km flight conducted in the final AERIAL-CORE demonstration.

\subsection{Multi-UAV Inspection with Autonomous Landing }
\label{sec::MultiUAVExperimentLanding}

When powerful storms hit an area, they usually generate foreign objects like trees and large plastic fabrics, falling to power lines and generating power outages. In this context, time is crucial to find the outage location and reestablish the power supply to citizens. To cope with that, AERIAL-CORE has developed a heterogeneous multi-UAV inspection system with cognitive capabilities, which consists of two DJI M210 multi-rotors and one DeltaQuad Pro fixed-wing aircraft with VTOL capabilities. All these robots have been endowed with autonomous navigation capabilities and are equipped with high-resolution visual cameras to capture and stream information from the environment.

\begin{figure*}[h!]
    \centering
    \includegraphics[width=1\textwidth]{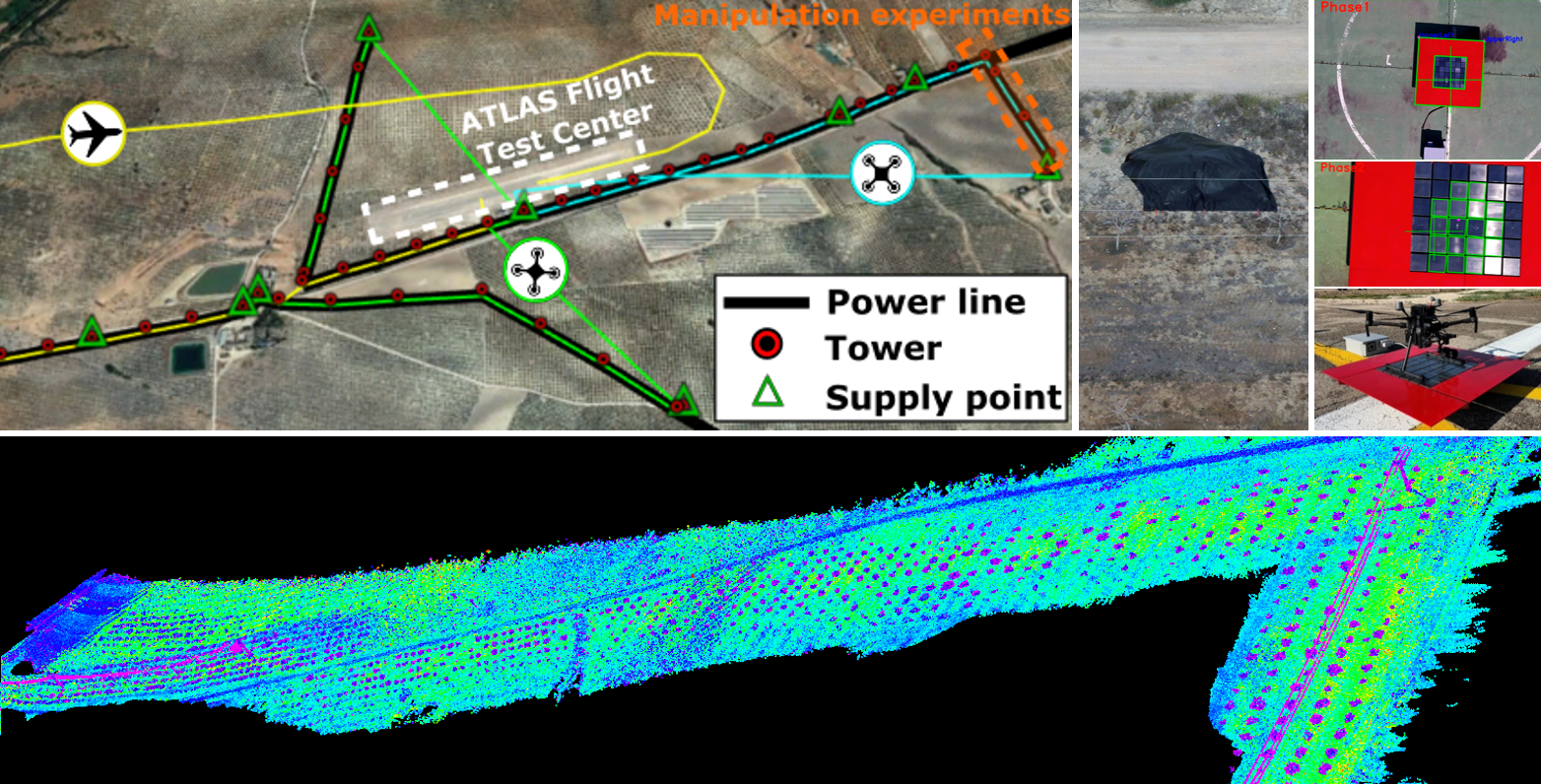} 
    \caption{ \textbf{(Top)} Autonomous inspection of ATLAS power grid by multi-UAV team: planned trajectories (top-left), a snapshot of the video streamed by one multi-rotor when a plastic foreign object hung from the line is detected (top-center) and autonomous landing on charging station (top-right). \textbf{(Bottom)} The resulting 3D map was obtained in the final AERIAL-CORE experiments.}
    \label{fig:multiUAV_inspection}
\end{figure*}

 A planning method for autonomous cooperative inspection of electric power grids was developed to achieve safe and efficient multi-UAV operation, minimizing the total inspection time. We exploit the heterogeneous capabilities of the UAVs and integrate terrain models for safe and accurate positioning and energy consumption models based on the UAV aerodynamics (including the wind effect and other weather conditions) to predict the optimal time to recharge the batteries. Moreover, due to the UAV's limited battery capacities, charging stations along the operation area are also considered within the planner for long-endurance operations. Whenever a UAV needs to land on a charging station, it uses visual servoing to enable accurate autonomous landing (see Figure \ref{fig:multiUAV_inspection} Top right). In the future, the charging stations using the magnetic field from the line (see Section IV-D) will be used.

The full system described above has been applied to the inspection of the power grid located in the surroundings of ATLAS (see Figure \ref{fig:multiUAV_inspection} Top left), demonstrating its capability to inspect autonomously and according to the end-user requirements of more than 10 km of power lines in 10 minutes.

\section{Aerial Manipulation for Maintenance}
\label{sec:AerialManipulation}

\subsection{Dual Arm Manipulation with Cart}
\label{sec::DAPC}

\begin{figure}[h!]
    \centering
    \includegraphics[width=1\columnwidth]{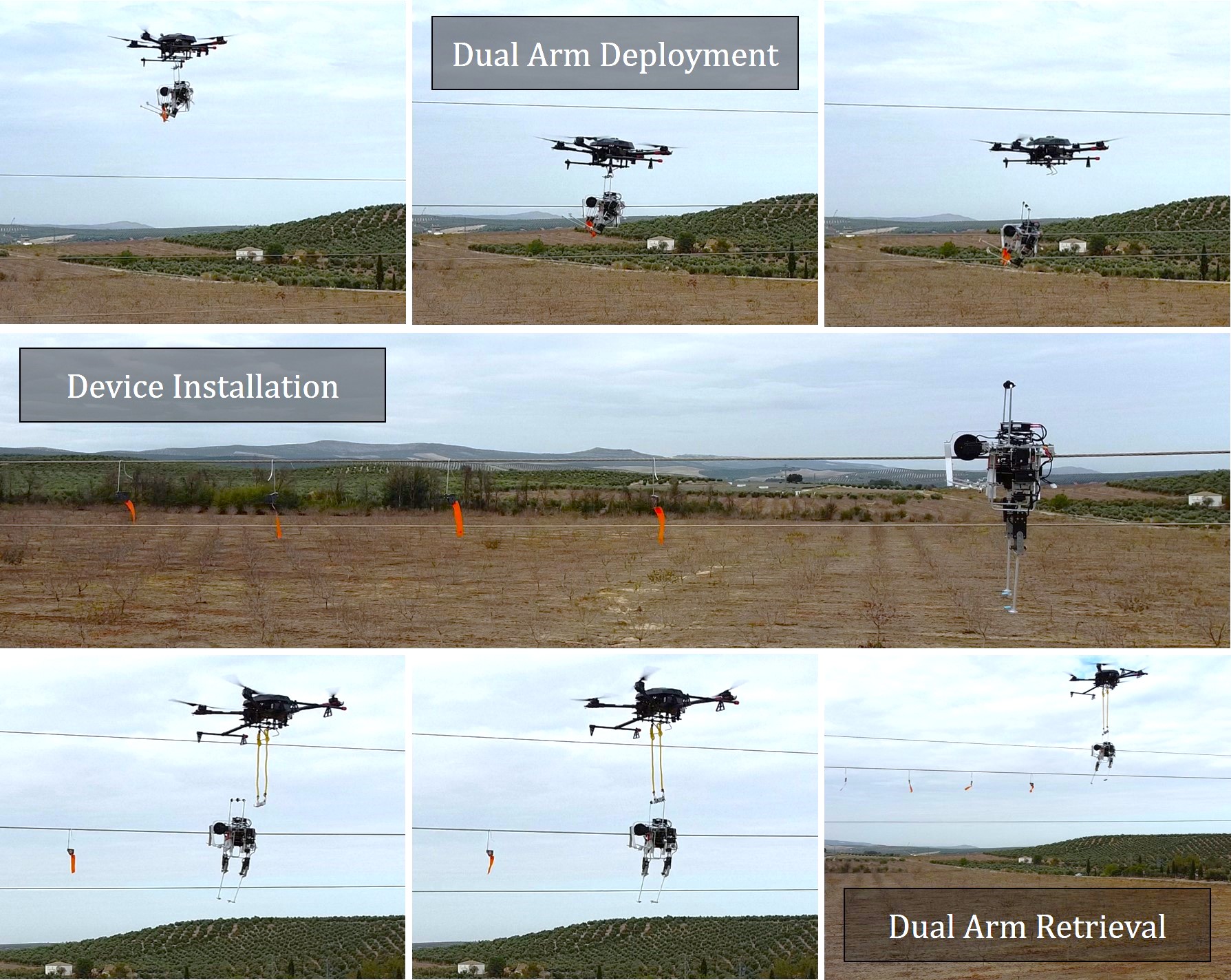}
    \caption{\textbf{(Top)} Aerial deployment. \textbf{(Middle)} devices installation. \textbf{(Bottom)} aerial retrieval of the dual arm rolling robot operating on the power line.}
    \label{fig:DAP-C_Experiment}
\end{figure}

Motivated by the benefits in terms of energy efficiency, accuracy, and reliability of performing the manipulation operations in perching conditions instead of doing it on flight \cite{suarez2022dual}, the dual arm platform with rolling base described in \cite{suarez2023ultra} has been applied in the installation of bird flight diverters on a real power line. The platform consists of a lightweight and compliant anthropomorphic dual-arm system LiCAS A1 (3 kg weight, 0.7 kg payload) equipped with a servo-driven rolling base (0.15 m/s speed) and magnetic grippers for grasping the devices \cite{suarez2023ultra}, using a quadrotor UAV (4 kg payload, 10 min flight time) with RTK-GPS for controlling the position and trajectory of the platform. The operation, illustrated in Figure \ref{fig:DAP-C_Experiment}, comprises three phases: 1) the aerial deployment of the dual arm robot on the power line using the multi-rotor, which is detached when a preloaded hook-handle mechanism is released once the arms are supported on the line, 2) the installation of the devices carried by the manipulator, and 3) the aerial retrieval using a double cable suspended magnetic hook mechanism. Each of the phases is executed in less than two minutes. Stable perching is achieved since the center of mass of the dual arm system is approximately 5 cm below the cable that supports the rolling base, similar to a pendulum, and thanks to the symmetric mass distribution of the arms. The device installation operation is programmed as a sequence of way-points for both end-effectors, detaching the device from the magnetic gripper by generating an impulsive motion against the cable, relying on the mechanical joint compliance of the arms to support the physical interaction.

\subsection{Main Local Manipulation Platform}
\label{sec::MLMP}

\begin{figure}[h!]
    \centering
    \includegraphics[width=1\columnwidth]{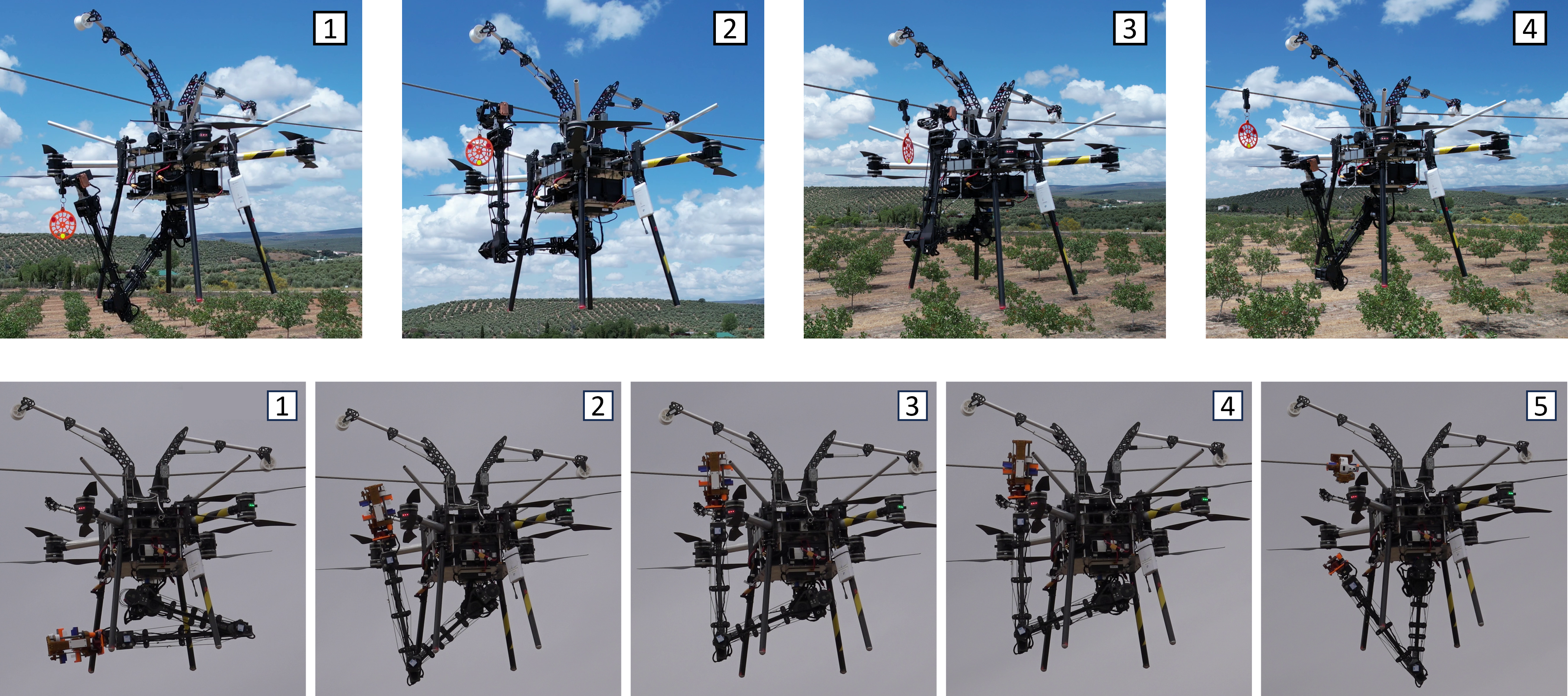}
    \caption{\textbf{(Top)} Clip-type bird diverter installation with MLMP. \textbf{(Bottom)} charging station installation with the same platform.}
    \label{fig:MLMP_validation}
\end{figure}

Rather than deploying a cart on the line and returning the UAV, the system's second platform was designed to explore a more general approach to the problem. The Main Local Manipulation Platform (MLMP) is an all-in-one solution, a single UAV that can fly to the line and detect it, perch autonomously on the line, and install or uninstall different devices as it moves along the cable. The voltage of the line does not need to be removed during the operation since the system has been designed and tested to handle voltages up to 125 kV because the grounded metal frame, which serves as a Faraday cage by protecting electronic components from electromagnetic interference while absorbing voltage arcs produced by the power line. The specially designed perching mechanism enables the system to perform the perching maneuver and to move along the power line easily and quickly by means of its integrated pulleys connected to DC motors. Additionally, the frame can integrate various tools and devices, such as the robotic arm that performs the manipulation. With a MTOW of 45 kg, MLMP is capable of flying for approximately 17 minutes, enough time to perform up to 5 perching maneuvers or a complete multi-device installation operation on a power line. 
MLMP was validated through many experiments (Figure \ref{fig:MLMP_validation}), including the installation and removal of a clip-type bird diverter and the installation of a charging station, which is  more complex and heavy. Both operations were performed multiple times on the actual power line with excellent results, demonstrating that MLMP  is a powerful and robust system and the first of its kind capable of autonomously perching on a power line, moving along it, and manipulating devices with a robotic arm.  

\subsection{Manipulator and End-Effector for the Main Platform }
\label{sec::ManipulatorEndEffector}

The robotic arm that is part of the MLMP has been designed explicitly for high payload manipulation. The manipulator is an anthropomorphic arm featuring six degrees of freedom, accompanied by a gear-based spherical wrist, to achieve a superior level of dexterity and enable the arm to accomplish all the designated tasks. The design of the arm discussed in detail in~\cite{marolla2022arm}, boasts an impressive payload of $5$~kg while weighing only $3$~kg; the remarkable payload/weight ratio of $1.67$ sets it apart from other similar solutions proposed in the literature. 

A different end-effector (EE) was mounted on the robotic arm for each validation experiment with the MLMP. For the installation of the clip-type bird diverters, a simple EE with one degree of freedom was made, which pushes the bird diverter mechanism until it closes (see Figure \ref {Figures/Figures_Section_IV/clip-teleop-charging} Top Left). For the installation of the small charging station, the EE with one degree of freedom was developed together with the charging station~\cite{stuhne2022design} (see Figure \ref {Figures/Figures_Section_IV/clip-teleop-charging} Bottom Left). This EE holds the charging station securely in place during transportation to the power line.  The charging station is attached to the power line and released from the EE  by actuating it. 

\subsection{Charging Stations and their Installation }
\label{sec::ChargingStation}

The magnetic field generated by the power line is the energy source for the charging station, which allows drones to recharge while inspecting high-voltage power lines. A very lightweight charging station has been developed. The station incorporates a split-core current transformer for energy harvesting and an electronic circuit to convert the harvested energy to DC for recharging the station's battery (see Figure \ref {Figures/Figures_Section_IV/clip-teleop-charging} Bottom Middle and Right). Further optimization of the design is achieved through the integration of the Transfer Window Alignment method with the Perturb and Observe algorithm and a Silicon Steel Core \cite{hoang2023advanced}. Experimental results revealed a significant 58.6\% increase in harvested power compared to traditional methods. Notably, our proposed approach allows the charging circuit to automatically identify the maximum power point, regardless of fluctuations in the current on the power lines, eliminating the need to sense the primary current.

A larger charger station capable of providing sufficient energy to charge heavy platforms was also developed.

\subsection{Teleoperation System }
\label{sec::Teleoperation}

\begin{figure}[h!]
    \centering
    \includegraphics[width=1\columnwidth]{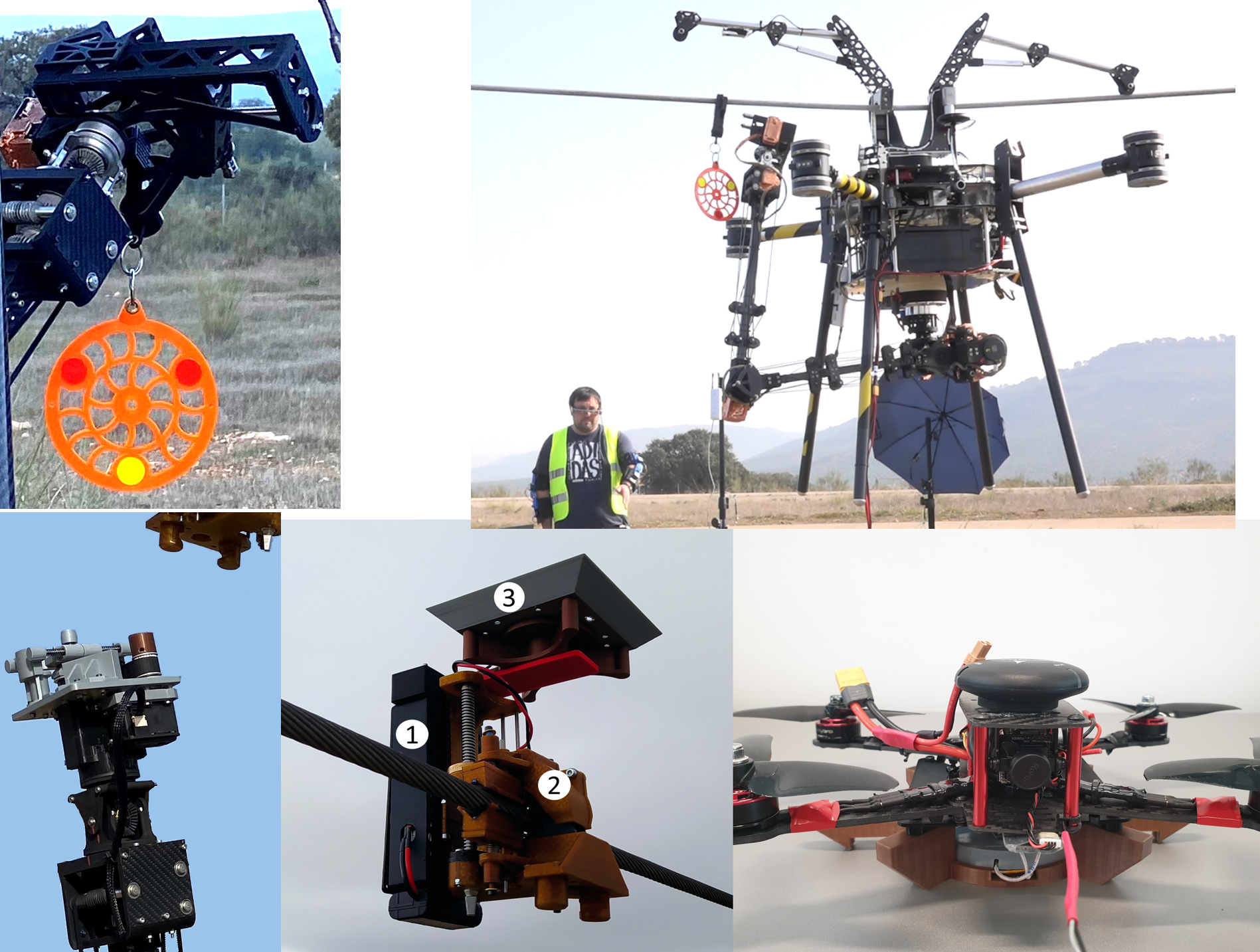}
    \caption{\textbf{(Top)} Left: Clip-type bird diverter end-effector.  Right: Installation of clip-type bird diverter with IMU-based teleoperation method. \textbf{(Bottom)} Left: Small recharging station end-effector. Middle:  The fully integrated system of the lightweight charging station (1 - the electronics compartment; 2 - split-core current transformer; 3 - V-shaped landing station). Right: The fully integrated system of the drone HolyBro QAV250 and the receiver assembly.}
    \label{Figures/Figures_Section_IV/clip-teleop-charging}
\end{figure}

In addition to the autonomous operation described in IV-B, a teleoperation system has been implemented. The IMU-based teleoperation method for the robotic arm on MLMP was validated under realistic conditions for installing clip-type bird diverters (Figure \ref{Figures/Figures_Section_IV/clip-teleop-charging} Top Right). The operator is equipped with 8 IMUs attached to different parts of his upper body, which in turn are used to estimate his upper body pose. The velocity references for each joint of the robotic arm are generated based on the operator's pose so that the operator's resting position gives a velocity of zero for all joints. Feedback from the platform to the operator is provided via a camera image projected onto the operator's smart glasses, allowing the operator to see the platform and simultaneously adopt a first-person perspective of the platform.

\section{Aerial Co-Working}
\label{sec:AerialCoWorking}

\subsection{Gesture Recognition}
\label{sec::GestureRecognition}

\begin{figure}
    \centering
    \includegraphics[width=1\columnwidth]{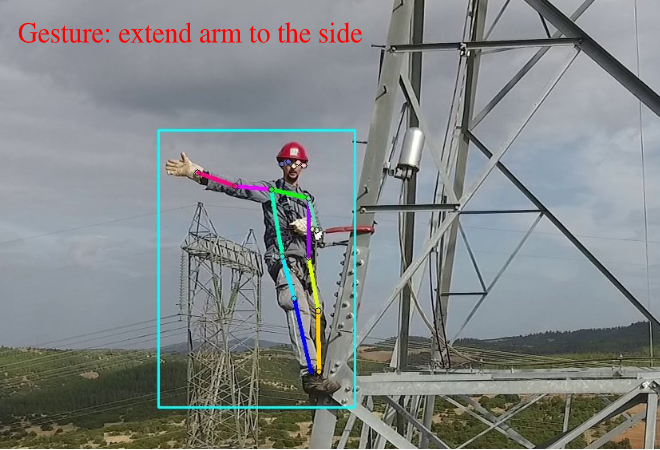}
    \includegraphics[width=1\columnwidth]{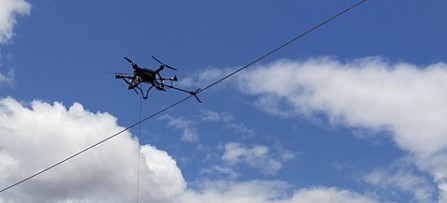}
    \caption{\textbf{(Top)} Visualization of all outputs (bounding box, 2D skeleton, predicted gesture) of the developed gesture recognition pipeline overlaid on the corresponding input video frame. \textbf{(Bottom)} Voltage check on ATLAS. The measured potential is compared to that of a ground cable to obtain the power line voltage, which is displayed on the ground station for the operator.}
    \label{fig:gestureDemo_contactDemo}
\end{figure}

In aerial co-working scenarios, human worker-UAV interaction and collaboration are critical. In this direction, we developed a human worker detection and gesture recognition system that can run smoothly onboard the UAV, enabling visual communication between human workers and UAV via gestures. First, RGB images from the onboard camera are processed on the fly for human detection and tracking. A fast deep neural object detector based on Single-Shot Detector (SSD) \cite{liu2016ssd} is employed in combination with a custom LDES-ODDA visual tracker. The output of this pipeline is a predicted bounding box for the tracked human in each input image where the human is visible, as depicted in Figure \ref{fig:gestureDemo_contactDemo}-Top. The predicted bounding boxes are then used in the gesture recognition pipeline. To maximize human worker detection accuracy, both the detector and the tracker were pre-trained on a manually annotated dataset\footnote{\url{https://aiia.csd.auth.gr/open-multidrone-datasets}} and then fine-tuned using videos of a human operator wearing safety equipment.

Given a sequence of images captured by an RGB camera of the UAV and the corresponding bounding boxes of the tracked human, the gesture recognition module recognizes the performed gestures from a predefined set (e.g., extend one arm to the side). 
The gesture recognition proceeds as a sequential pipeline. First, each video frame is cropped using the corresponding bounding box of the tracked human worker. Then, our method, as described in \cite{Papaioannidis2022tcsvt}, extracts a 2D human skeleton from each input image in pixel coordinates. The last $N$ outputs of the skeleton extractor, covering $N$ successive video frames, are stored in a FIFO buffer that is updated with a newly extracted 2D skeleton every $k$ video frame. This buffer is subsequently processed by our gesture classifier \cite{papaioannidis2021EUSIPCO}, a lightweight Long Short-Term Memory (LSTM) neural architecture that predicts the type of the performed gesture. The visualization of all the gesture recognition pipeline outputs for an example video frame can be seen in Figure \ref{fig:gestureDemo_contactDemo}-Top.
The pipeline was trained on a large, manually annotated dataset of gestures\footnote{\url{https://aiia.csd.auth.gr/auth-uav-gesture-dataset}}, and was further fine-tuned to perform effectively on aerial images. The parameters $N$, $k$ were empirically tuned to $N=9$ and $k=1$, based on the camera's update rate and the pipeline's performance when running onboard the UAV. Gesture recognition can also be used to command the voltage checking described below. 

\subsection{Aerial Co-Worker (ACW) Platform with  Interaction Control Laws and Voltage Checking}
\label{sec::ContactPlatform}
The ACW developed for co-working is a hexacopter with fixedly-tilted rotors, enabling full actuation and thus a precise, independent control of both its position and orientation, which is key when working close to humans. It has a 360° laser sensor, allowing it to navigate with centimeter precision in any environment with a known 3D model. The platform has been validated in two applications of  ATLAS: voltage check described below and tool delivery (Section V-C). For each use case, a specific payload was integrated into the hexacopter.

The voltage check application aims to measure the voltage on a power line to make sure it is safe for a human to carry out a maintenance task. To achieve this, a contact arm was integrated into the ACW, as depicted in Figure \ref{fig:gestureDemo_contactDemo}-Bottom. The tip’s shape was designed to facilitate contact with a conductor, and it is equipped with force and potential sensors.

To control the ACW's behavior during a physical interaction task, an impedance controller is designed. This control strategy modulates the robot's dynamics by imposing the desired inertia, damping, and stiffness on aerial robot behavior while in contact. This desired impedance is achieved using an admittance filter as an outer loop controller that provides reference signals to the internal loop motion controller. In particular, the admittance filter modifies the robot's reference trajectory based on the impedance parameters and sensed forces and torques, resulting in a new trajectory to be tracked by the aerial robot's internal motion controller. The choice of using an admittance filter in a hierarchical way with the motion controller allows for enhancing the motion controller for high-performance trajectory tracking in the presence of disturbances, such as wind.

\subsection{Tool Delivery}

Another application of the ACW is to deliver a tool to a worker operating in a hard-to-reach area, for example, while carrying out power line maintenance tasks. The worker guides the delivery using gestures (arm up, arms crossed, etc.) detected by the ACW applying the gesture recognition system described in Section \ref{sec::GestureRecognition}.
We integrated into the hexacopter a camera and Jetson AGX Xavier onboard computer to run the gesture recognition, as well as a delivery system consisting of a pulley and clamp.

\begin{figure}
    \centering
    \includegraphics[width=1\columnwidth]{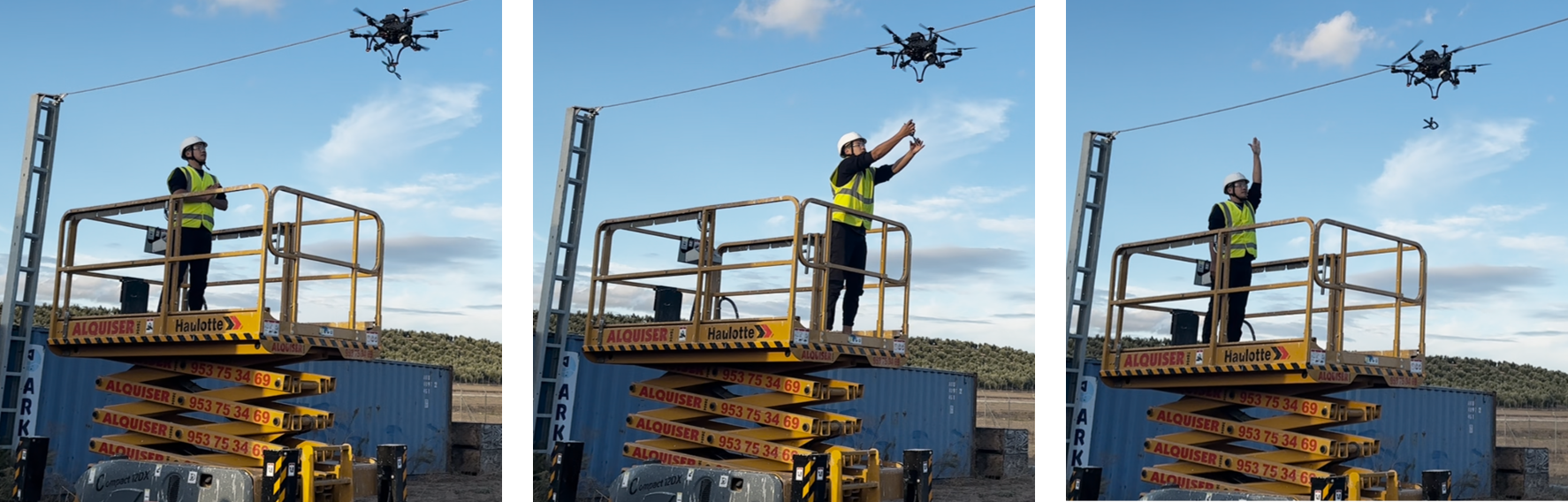}
    \caption{Tool delivery demonstration. \textbf{(Left)} The gesture "palms together" causes the pulley to descend. \textbf{(Middle)} The operator retrieves the tool. \textbf{(Right)} The gesture "arm up" commands the pulley to rise.}
    \label{fig:delivery-demo}
\end{figure}

The subsystem has been demonstrated successfully by handing out a tool to a worker on a lift near a line (see Figure \ref{fig:delivery-demo}).
The ACW flew to approximately 2~m above and 1~m next to the lift to ensure the worker's safety and the distance regulations.
Then, the worker gestured to the ACW to assist the delivery and retrieve the tool.

\subsection{Multi-UAV for Co-Worker Safety}
\label{sec::MultiUAVCoWorker}

The safety ACWs have been designed to monitor human operators during power line inspection and maintenance tasks while being able to offer the operator detailed additional, in-operation views of the power line infrastructure.
The UAV developed for this application is built on the Holybro X500 frame, with a MTOW of approximately 3.5 kg. It is equipped with a downward and upward-oriented depth camera and 3D LIDAR for obstacle detection and human operator localization, RTK GPS for UAV localization, and a front-facing RGB camera for power line inspection and monitoring the operator during the work.

In the monitoring phase, the formation of UAVs observes the human operator to ensure adherence to the safety rules of maintenance operations.
The leading UAV uses 3D LIDAR data to detect the human worker on the mobile lift platform.
Based on this detection, the UAV adjusts its position to provide the required view of the scene while additional UAVs maintain a formation defined in relation to the position of the monitored object and camera optical axis of the leading UAV, thereby offering additional perspectives of the monitored scene.
A multi-stage, MPC-based cooperative motion planning approach~\cite{kratky2021aerialFilming}, running onboard the UAVs, navigates the formation through the environment while ensuring collision avoidance and continuous monitoring of the worker.

\begin{figure*}[h!]
    \centering
    \includegraphics[width=1\textwidth]{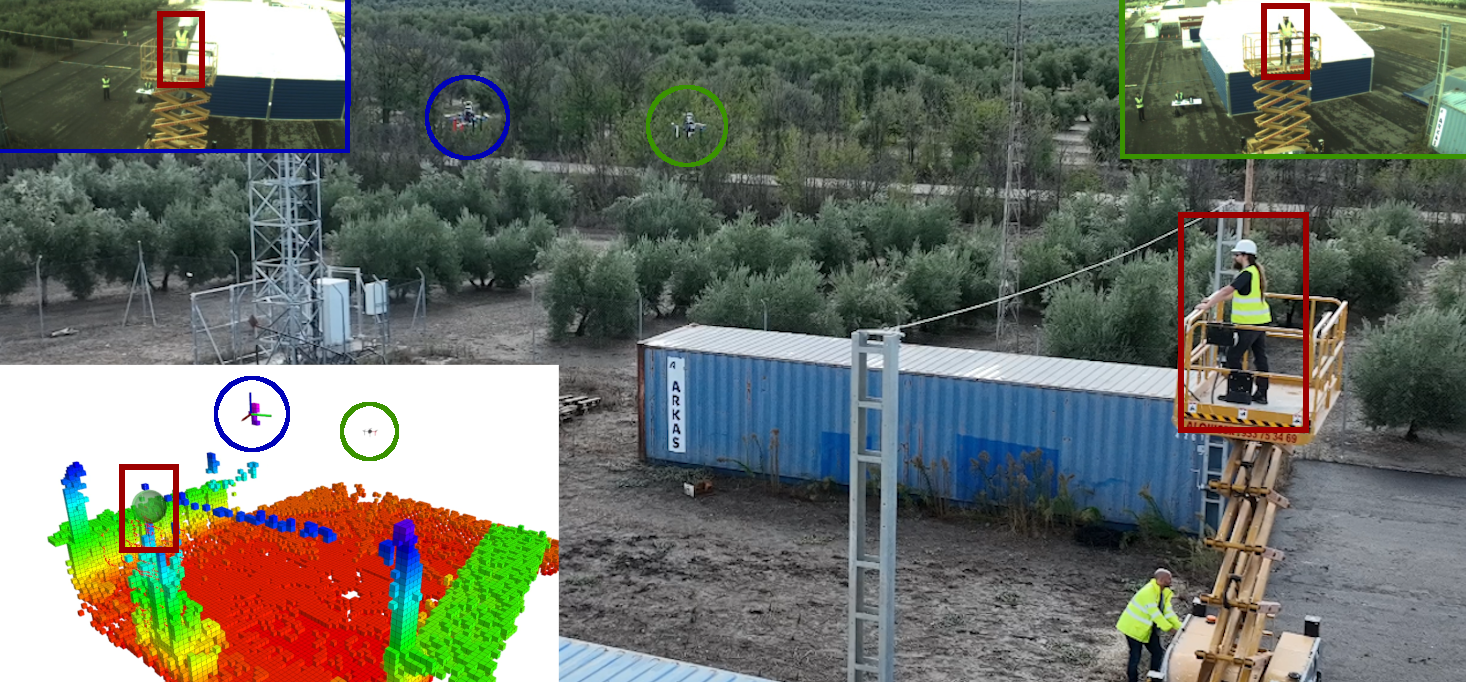}
    \caption{Snapshot showing safety monitoring of a human worker by a team of UAVs (highlighted by blue and green circles). In the top corners are the images from onboard cameras of particular UAVs, while the image in the bottom left corner shows the occupancy map built from the onboard LIDAR together with the detection of the human operator (green sphere in red rectangle).}
    \label{fig:acw_safety_demo_atlas}
\end{figure*}

The safety ACWs can serve as inspection ACWs in the same mission, assisting the operator in getting detailed data about the power lines from various viewpoints during ongoing maintenance operations. 
An arbitrary number of UAVs can be tasked to switch from safety monitoring to inspection mode. 
Utilizing the same motion planning approach~\cite{kratky2021aerialFilming}, the UAVs in inspection mode are navigated to the desired viewpoints while still avoiding collisions with other UAVs and obstacles. 
Once the inspection task is completed, the UAVs revert to monitoring mode and return to their designated safety monitoring positions.
The whole system, relying fully on onboard computation, was validated through numerous experiments in scenarios, including both monitoring and inspection tasks.
A snapshot from the real-world validation is shown in Figure~\ref{fig:acw_safety_demo_atlas}.

\section{Integrated AERIAL-CORE System }
\label{sec:IntegratedSystem}

The integrated AERIAL-CORE system is the application managing the operation logistics of the three subsystems presented in Sections III, IV, and V. The process  consists of the following steps:  a) The Chief Inspector (CI) receives instructions from the utility (electric power distribution) company (1 in Figure \ref{fig:/Figures_Section_II/aerial-core_system}); b) The CI starts with the long-range inspection using the integrated AERIAL-CORE system by including all the necessary information in it; c) The inspection  team 1, consisting of  multiple workers and robots, receives the mission; d) The team performs the long-range inspection (2 in Figure \ref{fig:/Figures_Section_II/aerial-core_system})  by  using the methods and technologies in section III and the workers send feedback information to the CI by using the AERIAL-CORE system;  e) Once the inspection is completed, the workers take the results (report 3 in Figure \ref{fig:/Figures_Section_II/aerial-core_system}) and send them to the CI to be analyzed; f) The CI activates  the manipulation team 2 (4 in Figure \ref{fig:/Figures_Section_II/aerial-core_system}) and the co-working team 3 (6 in Figure \ref{fig:/Figures_Section_II/aerial-core_system}); g) The co-working team starts the operation by checking if the power line has voltage or not following Section V and the results are also sent to the CI; h) The manipulation team performs the installation following Section IV, and the results (report 5 in Figure \ref{fig:/Figures_Section_II/aerial-core_system}) are sent to the CI;  i) A worker needs to do a manual operation by climbing the tower, or using an elevation platform; j)  Once there, he/she notices that forgets one tool and then calls to the co-worker robot to bring it to him/her following Section V; k) The foreign object is removed and a report (7 in Figure \ref{fig:/Figures_Section_II/aerial-core_system}) is sent to the chief inspector. Finally,  the CI sends the report (8 in Figure \ref{fig:/Figures_Section_II/aerial-core_system}) to the electric power distribution company.

All the operations were performed successfully in a real-time demonstration on October 27, 2023.

\section{Conclusion}
\label{sec:Conclusion}
The relevance and interest of intelligent aerial robotics systems and their application to I\&M of electrical power distribution systems have been demonstrated for the first time in this paper. Multiple heterogeneous intelligent aerial robots flying BVLOS under European (EASA) and national civil regulations can cooperatively inspect hundreds of kilometers of power lines. They can perform VTOL and complete detailed inspections at low velocity or in hovering and morph to minimize energy consumption during the long-range inspection. They can perform autonomous tracking of the lines and mapping of the vegetation near the lines, which is very important to avoid forest fires. They can also perch on the lines to recharge the battery or to realize maintenance tasks by means of robotic manipulators, including the installation of bird diverters to protect natural life.  The aerial robots can also help human workers at high altitudes by measuring the voltage of the line, providing them with tools and monitoring their safety. A video with a summary of the AERIAL-CORE project can be found at \url{https://www.youtube.com/watch?v=Oyw7VwM7sCs}. The expected impact of this work is very high because the I\&M of electrical power lines has a cost of many billions in the world (more than 2.2 billion euros only in Europe), and there is also a direct impact on the safety of the workers. 

The work presented in this paper can also open new paths, such as implementing drone highways over the lines, with the robots perching on the line to charge batteries.

\section{Acknowledgment}
\label{sec:Acknowledgement}

This work was supported by the EU Horizon 2020 Research and Innovation Program under grant agreement no. 871479.

\bibliographystyle{IEEEtran}
\bibliography{Main.bib}

\end{document}